\documentclass[conference,letterpaper]
{IEEEtran}
\usepackage[utf8]{inputenc}
\usepackage[T1]{fontenc}
\usepackage[protrusion=false,expansion=false]{microtype}
\usepackage{booktabs}
\usepackage{multicol}
\usepackage{newtxtext}
\usepackage{nohyperref}
\PassOptionsToPackage{hyphens}{url}
\usepackage{url}
\usepackage{amsmath,amsfonts,amssymb}
\usepackage{graphicx}
\usepackage{tabularx,array}
\usepackage{xcolor}
\usepackage[table]{xcolor}
\usepackage{nicefrac}
\usepackage[caption=false,font=footnotesize]{subfig}
\usepackage{cite}

\providecommand{\AND}{\and}

\title{Data-driven Lake Water Quality Forecasting for Time Series with Missing Data using Machine Learning}

\author{Rishit Chatterjee\\Department of Computer Science, Colby College\\
\texttt{rchatt28@colby.edu}\\
\AND
 Tahiya Chowdhury\\Department of Computer Science, Colby College\\
 \texttt{tahiya.chowdhury@colby.edu}\\
}

\begin{document}
\raggedbottom
\maketitle

\begin{abstract}

Volunteer-led lake monitoring yields irregular, seasonal time series with many gaps—arising from ice cover, weather-related access constraints, and occasional human errors—complicating forecasting and early warning of harmful algal blooms. We study Secchi Disk Depth (SDD) forecasting on a 30-lake, data-rich subset drawn from three decades of in-situ records collected across Maine lakes. Missingness is handled via Multiple Imputation by Chained Equations (MICE), and we evaluate performance with a normalized Mean Absolute Error (nMAE) metric for cross-lake comparability. Among six candidates, ridge regression provides the best mean test performance. Using ridge regression, we then quantify the minimal sample size, showing that under a backward, recent-history protocol, the model reaches within 5\% of full-history accuracy with \(\approx 176\) training samples per lake on average. We also identify a minimal feature set, where a compact four-feature subset matches the thirteen-feature baseline within the same 5\% tolerance. Bringing these results together, we introduce a joint feasibility function that identifies the minimal training history and fewest predictors sufficient to achieve the target of staying within 5\% of the complete-history, full-feature baseline. In our study, meeting the 5\% accuracy target required about 64 recent samples and just one predictor per lake, highlighting the practicality of targeted monitoring. Hence, our joint feasibility strategy unifies recent-history length and feature choice under a fixed accuracy target, yielding a simple, efficient rule for setting sampling effort and measurement priorities for lake researchers. 

\end{abstract}


\section{Introduction}
\label{sec:intro}

\subsection{Background}
Throughout the world, lakes are vital for drinking-water security, biodiversity, recreation, and regional economies, storing the majority of accessible surface freshwater~\cite{Liquete2011}. Yet, water quality is declining under global warming, altered stratification, and nutrient enrichment, heightening the risk of harmful algal blooms (HABs)~\cite{Anderson2012}. Blooms are propelled by over-nutrient (especially phosphorus) and stratification, and reduce transparency, deplete oxygen, and can produce cyanotoxins that jeopardize drinking-water safety and ecosystem health~\cite{Sukenik2021, Smucker2021}.
Effective lake management requires timely monitoring of water clarity~\cite{Wang2018}, often achieved by citizen-led efforts. While citizen-collected data plays a critical role in tracking water clarity, these volunteer-led monitoring programs yield irregular, seasonally intermittent time series with missing data due to ice cover, access constraints, sensor outages, and storm events—compromising trend estimation, forecasting, and early warning~\cite{Rand2022, Heddam2016}. In practice, sampling is often skipped during adverse weather, and many lakes are visited too infrequently to provide consistent records. Moreover, in a changing climate, lakes are increasingly affected by shifts in precipitation, warming temperatures, and changing nutrient cycles. Monitoring and forecasting water clarity is therefore crucial not just for ecosystem health, but for anticipating climate-exacerbated algal bloom events and informing mitigation strategies.

\subsection{Related Work}
Prior work on lake water-quality monitoring has predicted Secchi Disk Depth (SDD) and bloom risk using satellite reflectance and in-situ chemistry--based machine learning (ML). For example, \cite{Zhang2022secchi} derived optical water-penetration properties from spectral indices to estimate SDD in five lakes, and \cite{Izadi2021} leveraged high-resolution hyperspectral imagery to provide early warnings of algal blooms. Remote sensing can mitigate data sparsity resulting from sampling, but requires high-resolution imagery and regional calibration, which limits deployment~\cite{Olmanson2015}. Moreover, ~\cite{Sagan2020ESR} concludes that current sensors and estimation methods have important limitations (e.g., stringent atmospheric correction and ground-truth requirements), hindering generalizable water-quality products at the scales of small inland (i.e., water bodies surrounded by land and not part of coastal waters) lakes, as in our study \cite{Palmer2015InlandWaters}. In ~\cite{Hanly2025LAGOSUS}, lakes $< 4\ \text{hectares (ha)}$ are excluded because smaller waterbodies produce mixed water--land pixels that prevent water-leaving reflectance from optically deep water. Even for $\geq 4$\,ha lakes, scenes must be masked for clouds/snow/ice and, on average, only $\approx$~48\% of pixels per lake--scenes are retained, with SDD predictions explaining at most $\approx$~63.7\% of variance and scene availability shaped by regional cloud patterns. Anand et al.~\cite{Anand2024EcolInf} used ML with multispectral imagery for water-quality prediction, but it depends on sensor-specific trade-offs (spectral vs. spatial) and cloud-free scenes — conditions rarely met across Maine’s many small, shoreline-complex lakes. Hence, given our many small, narrow-shoreline, seasonally cloudy/icy Maine lakes, these constraints make remote sensing-based water quality monitoring unreliable. 

Prior in-situ studies have attempted to model manually sampled nutrients and physicochemical variables—such as temperature, phosphorus, nitrogen, and dissolved oxygen—to predict chlorophyll and clarity with machine-learning methods~\cite{Mamun2020}, and neural networks to estimate SDD~\cite{Kulisz2021}. Despite having advanced prediction capability, most of these studies were evaluated on only a small number of lakes, limiting generalizability. Moreover, while predicting SDD from in-situ chemistry is common, handling sparsity and seasonally structured missingness remains underexplored. Prior work often handles data gaps via complete-case (or listwise) deletion, which is valid only for missing completely at random (MCAR) values. In hydrologic and environmental time-series forecasting, missingness—if handled by deletion—can degrade predictive performance~\cite{Rodriguez2021, Meisenbacher2022, Gill2007, Pedersen2017}. Additionally, many works evaluate forecasting performance using Mean Absolute Error (MAE) and Root Mean Square Error (RMSE)~\cite{He2024} that are scale-dependent and impede cross-lake comparability. 

Several recent efforts also examine how \textbf{training sample size} affects clarity measurements, especially in remote sensing of SDD. For global lakes, \cite{Zhang2022SDD_ML} reports that model accuracy is sensitive to the number of matched in-situ observations used to train machine-learning regressors, with analyses of performance vs.\ training set size. Large-scale datasets such as LAGOS-US LANDSAT further emphasize data quality control and provide variance-explained benchmarks for SDD, bringing forth practical constraints (cloud masking, pixel availability) that limit usable samples per lake and date \cite{Hanly2025LAGOSUSLandsat}. When it comes to the \textbf{number of features} needed, we find band/ratio selection and model-driven importance ranking to be common. Reviews and frameworks document feature selection for water-quality retrieval, and empirical SDD studies often identify a small set of informative bands/ratios (e.g., blue/red, blue/green) using variable importance in random-forest models. \cite{Rubin2021LakeClarity, Pang2025DLInlandWQReview, Hanly2025LAGOSUSLandsat, Yao2025HybridAttentionWQ}. Related efforts also attempt to \textbf{connect training data sufficiency and feature selection}. Cross-basin work shows that deep representation learning can retain strong predictive skill when trained on reduced data sizes and also explores exogenous meteorological drivers with attention mechanisms \cite{Zheng2025CrossBasinWQ}. Thus, there is a need for understanding the role of sample size (the number of samples) and the optimal features required for forecasting water quality across numerous lakes. However, these approaches either vary the amount of training history or identify compact predictor sets. They stop short of a unified selection rule that simultaneously balances recent-history length and the feature set against a fixed accuracy target across lakes. This gap motivates our study.

\subsection{Contribution}
To address irregular, seasonally intermittent records and heterogeneous data across lakes, we propose a practical forecasting framework that determines how much recent history is sufficient and which measurements are essential for reliable SDD prediction. We then consolidate these decisions with a single selection rule that balances training-sample size and the feature set against a fixed accuracy target. The primary contributions of our research are listed as follows:

\begin{itemize}
  \item \textit{Minimal samples.} Using a backward (recent-history) training protocol, we quantify the smallest training data size needed to meet a fixed accuracy tolerance: models reach within 5\% of a complete-record, full-feature reference with \(\approx 176\) training samples per lake on average, in line with learning-curve–based sample-size planning \cite{Dayimu2024LearningTypeCurves}.
  \item \textit{Minimal features.} We identify a compact predictor subset by ranking features based on Mean Decrease in Impurity (MDI) and performing forward selection with the ridge forecaster; a four-feature set matches the thirteen-feature baseline within the same 5\% tolerance, reducing measurement burden.
  \item \textit{Unified selection rule.} We introduce a single feasibility rule that simultaneously chooses the shortest recent history and smallest predictor set that still meets an accuracy target of staying within 5\% of a reference model developed from all available training samples and the full predictor set.
\end{itemize}



\begin{table}[t]
\caption{Field measurements listed by their percentage gaps in the dataset. We excluded chlorophyll and "SDD-to-bottom" observations}
\label{tab:vars}
\centering
\footnotesize
\begingroup
\setlength{\tabcolsep}{2pt}            
\renewcommand{\arraystretch}{1.12}     
\newcolumntype{Y}{>{\raggedright\arraybackslash}X} 
\begin{tabularx}{\columnwidth}{@{}c l Y r@{}}
\toprule
\textbf{\#} & \textbf{Data} & \textbf{Description (unit)} & \textbf{\% gap} \\
\midrule
1  & \texttt{midas}             & MIDAS lake code                                & --   \\
2  & \textit{lake}              & Lake name                                      & --   \\
3  & \textit{max depth}         & Lake depth (m)                                 & 1.35 \\
4  & \textit{surface oxygen}    & Dissolved O\textsubscript{2} at surface (mg/L) & 74.32 \\
5  & \textit{Schmidt stability} & Energy to fully mix lake (J/m\textsuperscript{2}) & 71.81 \\
6  & \texttt{TSc}               & Surface temperature (\(^{\circ}\)C); mean where depth \(\le\) 1 m & 70.20 \\
7  & \texttt{zTm}               & Thermocline depth (m); first depth with \(dT/dz \ge 1^{\circ}\mathrm{C}/\mathrm{m}\) & 60.10 \\
8  & \texttt{TBc}               & Bottom temperature (\(^{\circ}\)C)               & 71.20 \\
9  & \texttt{P\_sp\_ppb}        & Surface phosphorus (ppb) from epicore samples   & 79.10 \\
10 & \texttt{P\_b\_ppb}         & Bottom phosphorus (ppb) from bottom grabs       & 92.51 \\
11 & \texttt{zS\_m}             & Secchi disk transparency (m)                    & -- \\
\bottomrule
\end{tabularx}
\endgroup
\end{table}

\section{Methods}
\label{sec:methods}

\subsection{Data}
Our dataset consists of lake monitoring records for Maine spanning three decades. We create this dataset by merging monitoring record archives from the Maine Department of Environmental Protection (MDEP) and Lake Stewards of Maine via \texttt{MIDAS IDs}, yielding $793$ lakes with irregularly sampled, human-collected time series. This dataset includes field observations (temperature, dissolved oxygen, SDD), laboratory measurements (nutrients such as total phosphorus; chlorophyll-\textit{a}), and derived lake physics (mixed-layer depth, oxic status, Schmidt stability from \texttt{rLakeAnalyzer}) which form a lake-specific multivariate time series (Table ~\ref{tab:vars}). 

After consulting with domain experts in lake water conservation, we decided to exclude chlorophyll to prevent target leakage as chlorophyll closely governs SDD transparency and acts as a direct proxy. We also drop “SDD-to-bottom” observations (SECCBOT = "Yes") as non-determinable measurements (if the disk reaches the bottom and is still visible, the SDD cannot be determined). After this step, we select the top 30 lakes with the most observations, ensuring adequate temporal coverage for imputation and credible time-series evaluation.

\textit{How do we select the top 30 lakes?} We let $t\in\{1,\dots,T\}$ index timestamps and $j\in\{1,\dots,p\}$ index the $p$ predictors.
Denote by $z_{tj}$ the value of predictor $j$ at time $t$. “NA” indicates a missing entry.

For each predictor $j$, define the missingness rate as
\[
m_j \;=\; \frac{1}{T}\,\bigl|\{\,t\in\{1,\dots,T\} : z_{tj}\ \text{is NA}\,\}\bigr|.
\]

The lake-level predictor missingness is
\[
\bar m \;=\; \frac{1}{p}\sum_{j=1}^{p} m_j,
\]
and we rank lakes by the amount of data $1-\bar m$, selecting the 30 lakes with the smallest $\bar m$ for further analysis. As a feasibility-oriented proof-of-concept, we use this information-rich subset to validate further experiments (see Appendix~\ref{app:tt} for the list of 30 lakes).

Then, for \textbf{forecasting} purposes, we treat SDD as the target and use the remaining physicochemical, nutrient, and stratification features as predictors.


\subsection{Time-series Imputation}
Let $L\in\{1,\dots,\ell\}$ index lakes, and let
$\{t_k\}_{k=1}^{T_\ell}$ be the observation times for lake $\ell$, where
$T_\ell$ is the number of timestamps available for that lake.
At each $t_k$ we record the SDD as $y_{\ell,t_k}$ and a feature vector
$\mathbf{z}_{\ell,t_k}\in\mathbb{R}^{\,d-1}$, where $d$ is the total number of variables
(SDD $+$ other features).

\noindent By stacking features over time, we get the \emph{covariate series} as,
\[
Z_\ell \;=\;
\begin{bmatrix}
\mathbf{z}_{\ell,t_1}\\[-2pt]
\vdots\\[-2pt]
\mathbf{z}_{\ell,t_{T_\ell}}
\end{bmatrix}
\in \mathbb{R}^{\,T_\ell \times (d-1)},
\]
with observed and missing entries denoted by $Z_{\ell,\mathrm{obs}}$ and
$Z_{\ell,\mathrm{miss}}$, respectively. Before forecasting future clarity
$y_{\ell,t+h}$ at horizon $h$, we complete the covariate series by
imputation.

Prior works have explored statistical imputation using mean and zero~\cite{nijman2022missing}, regression~\cite{afrifa2020missing}, maximum likelihood, expectation maximization~\cite{dempster1977maximum}, multiple imputation~\cite{rubin1987multiple} techniques for handling missing data in multivariate continuous-valued time series cases. More recently, neural network-based techniques have been proposed to handle time series missing data with network augmentation~\cite{che2018recurrent} and data generation~\cite{10.1007/s00521-022-08165-6}. Based on our preliminary experiments, we selected Multivariate Imputation technique with Chained Equation (MICE)~\cite{van2011mice}, where each variable with missing data is modeled conditional upon the other variables available in the data. We chose \textit{MICE} over other similar techniques such as MissFOREST~\cite{stekhoven2012missforest}, as \textit{MICE} can handle both continuous and categorical values, and works under the assumption that data is missing at random. We also note that while imputation methods that fill missing data using deep generative networks~\cite{muzellec2020missing, Lall_Robinson_2022, mattei2019miwae, kyono2021miracle, yoon2018gain} require no distributional assumptions, the increased computational cost does not provide significant performance improvement, as we observed in our analysis.

Based on these observations, we use \textit{Multiple Imputation by Chained Equations (MICE)}
under fully conditional specification~\cite{Mbuvha2022, Mahmood2024, Hamzah2022}. Each variable with gaps is
modeled conditionally on the others and iteratively imputed to produce a
\emph{completed covariate matrix}, $\tilde Z_\ell\in\mathbb{R}^{\,T_\ell\times(d-1)}$. This provides us the imputed matrix for $\{t_k\}$ with no
missing entries. It is also to be noted that \textbf{SDD is never imputed} as it is the target variable. We use this completed data matrix $\tilde Z_\ell$ (together with targets $\{y_{\ell,t_k}\}$) then in the backward-forecasting step explained next.

\subsection{Minimal Sample Size with Backward Forecasting Strategy}

\noindent \textbf{Why is determining the minimal number of samples important?} Field sampling is labor- and resource-intensive and irregularly sampled across lakes and seasons. Thus, it is important to estimate the \emph{smallest lake data history}. Beyond this sample size, further sampling achieves no further gains in forecasting performance and can improve resource allocation.

Traditional time-series forecasting models use the earliest available observations and utilize them to predict future measurements~\cite{Tashman2000OOS}. However, this strategy does not inform the forecasting model of the most recent data patterns, which are particularly relevant to capturing the changing climate and weather patterns. We evaluate forecasting under a backward-expanding protocol: for each lake $\ell$, we hold out the most recent test block (e.g., last 5 years) with timestamp set $\mathcal{T}^{\mathrm{test}}_\ell$ and use the remaining pre-test history as the candidate training pool. In this strategy, our data is evaluated on the most recent 5 years' data and the training set contains the remaining data prior to those 5 years (pre-test). We choose a five-year test block because it spans multiple seasonal cycles while emphasizing recent conditions, and it aligns with commonly used five-year assessment/review cycles in water-quality programs \cite{Tashman2000OOS, Green2025WRTDS}.

Let $N_\ell^{\mathrm{pre}}$ be the number of pre-test samples after chronological sorting, and let $\mathcal{N}_\ell=\{n_{\min},\dots,N_\ell^{\mathrm{pre}}\}$ be the grid of training sizes (with $n_{\min}$ the smallest size that permits model fitting). 
\noindent For each $n\in\mathcal{N}_\ell$, we first train on the most recent $n$ pre-test samples and predict all $t\in\mathcal{T}^{\mathrm{test}}_\ell$.


\subsection{Evaluation Metric: Normalized MAE}

We report the forecasting performance with the \emph{normalized MAE (nMAE)} metric, which is defined as:
\[
\mathrm{nMAE}_\ell(n)
=\frac{1}{\lvert \mathcal{T}^{\mathrm{test}}_\ell\rvert\,\overline{y}_{\ell,\mathrm{test}}}
\sum_{t\in\mathcal{T}^{\mathrm{test}}_\ell}\bigl|\,y_{\ell,t}-\hat y_{\ell,t}^{(n)}\,\bigr|,
\]
\emph{nMAE} is the average absolute error on the test block divided by the \emph{average} SDD on that block, which makes errors comparable across lakes that have different average clarity and depth. This scale normalization is particularly useful for climate-relevant water-quality modeling problems across diverse geographies, where lakes can differ in baseline clarity due to climate and land-use effects. nMAE has been shown to be useful previously in benchmarking point forecasts across diverse horizons \cite{Zhang2024ProbTS}. It has also been used previously in time-series forecasting problems as a scale-comparable metric \cite{AlShafeey2024Heliyon}. In addition, we also report MAE and $R^{2}$, both widely used accuracy measures for forecasting \cite{Chicco2021R2}.

Let $\mathrm{nMAE}_\ell^{\mathrm{final}}=\mathrm{nMAE}_\ell(N_\ell^{\mathrm{pre}})$ be the reference using all pre-test samples. Then the \emph{minimal sample count} becomes,
\[
n_\ell^\star=\min\Bigl\{n\in\mathcal{N}_\ell:\ \mathrm{nMAE}_\ell(n)\le 1.05\cdot \mathrm{nMAE}_\ell^{\mathrm{final}}\Bigr\},
\]
so that additional samples yield $\le$5\% improvement compared to that with complete data history. We use 5\% as a diminishing-returns tolerance (near-best performance), adjustable to reflect accuracy--cost trade-offs~\cite{Koshute2021TrainingSetSizes,Meek2002LearningCurveSampling}.



\subsection{Feature Selection By Importance Ranking}
\label{sec:feature_ranking}

\noindent \textbf{Why is determining the minimal number of features important?} Not all features are equally informative, and many are costly to measure. Ranking allows us to quantify a feature's contribution to the modeling process, reduce measurement/computation cost, and improve interpretability under collinearity ~\cite{Jakeman1993}. 

\begin{figure*}[!t]
  \centering
  \includegraphics[width=0.90\linewidth]{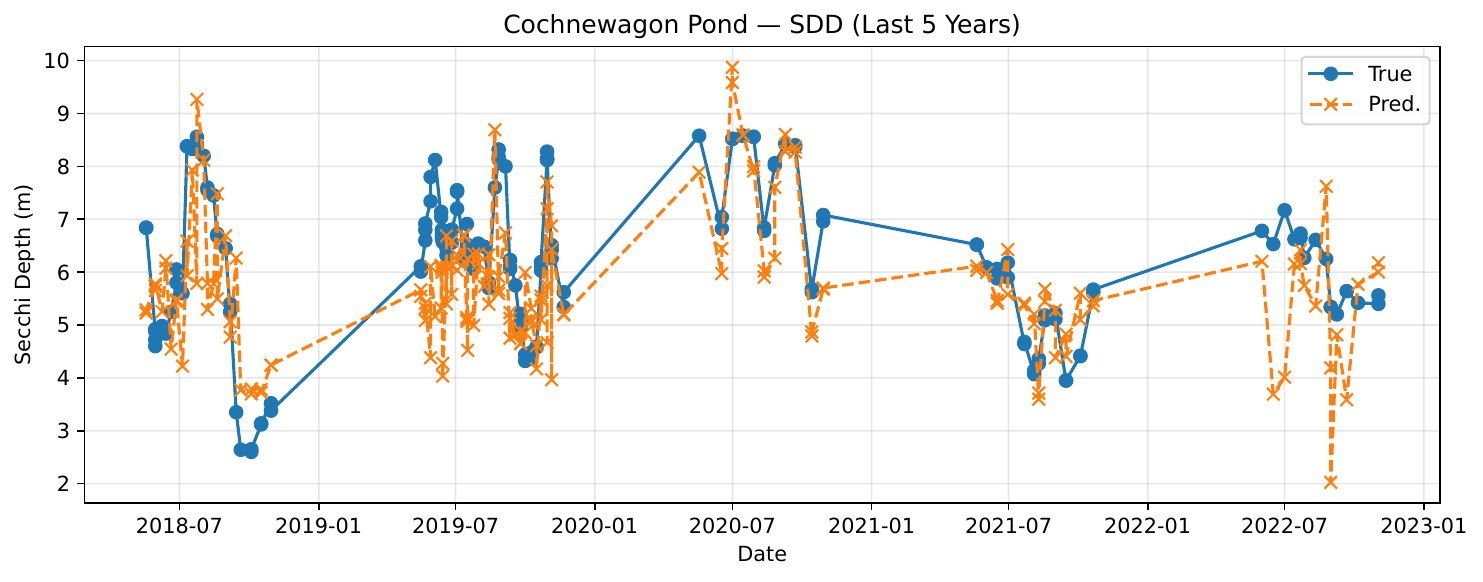}
  \caption{Cochnewagon Pond: observed and ridge-predicted SDD over the held-out last five years. The x-axis represents chronological sampling dates and the y-axis is the SDD (m). Observed values are shown as solid circles; ridge predictions (after MICE covariate imputation) are shown as dashed crosses, illustrating seasonal variability and agreement over the test period.}
  \label{fig:cochnewagon-sdd-ridge}
\end{figure*}

We evaluated six candidate forecasting techniques(Table~\ref{tab:model-means}) in this work. After being trained on all 13 features, ridge regression performed best in the forecasting task as it achieved the lowest MAE, nMAE, and also the least-negative $R^{2}$ (see Appendix~\ref{app:tt}, Table~\ref{tab:ridge-metrics}, for a comparison of train and test errors).  Fig.~\ref{fig:cochnewagon-sdd-ridge} illustrates the ridge regression forecasts over the test period as a time series. 

It also outperformed the classical statistical time-series baselines (ARIMA/SARIMA) \cite{Hyndman2008ForecastJSS, Szostek2024Energies}, in addition to our chosen deep-learning models. So we use it as the forecaster for the remainder of the paper and in the results section.

For the purposes of these experiments, ARIMA and SARIMA were fit with \texttt{statsmodels} \cite{Seabold2010Statsmodels}. Deep learning models (\texttt{TSMixerModel} and \texttt{TransformerModel}) were trained with \texttt{Darts} \cite{Herzen2022Darts}. Finally, \texttt{Ridge} and \texttt{RandomForestRegressor} were implemented with \texttt{scikit-learn} as supervised regressors, following the library’s standard \texttt{fit}/\texttt{predict} workflow \cite{scikit-learn}. 

\begin{table}[t]
\centering
\caption{Pre- vs.\ post-imputation mean test performance across 30 lakes. \emph{Unimputed}: complete-case after dropping chlorophyll and SECCBOT{=}Yes. \emph{Imputed}: MICE-completed covariates. We observe that imputation improves performance across all models.}
\label{tab:model-means}

\scriptsize
\setlength{\tabcolsep}{3pt}
\renewcommand{\arraystretch}{1.08}

\begin{tabular}{@{}l rrr rrr@{}}
\toprule
& \multicolumn{3}{c}{Unimputed} & \multicolumn{3}{c}{Imputed} \\
\cmidrule(lr){2-4}\cmidrule(lr){5-7}
\textbf{Model} & \textbf{MAE} & \textbf{nMAE} & \textbf{R2} & \textbf{MAE} & \textbf{nMAE} & \textbf{R2} \\
\midrule
ARIMA         & 0.950 & 0.185 & $-0.450$ & 0.883 & 0.169 & $-0.401$ \\
SARIMA        & 1.000 & 0.195 & $-0.630$ & 0.921 & 0.180 & $-0.577$ \\
TSMixer       & 0.820 & 0.225 & $-0.640$ & 0.748 & 0.204 & $-0.578$ \\
Transformer   & 0.900 & 0.245 & $-0.760$ & 0.829 & 0.223 & $-0.699$ \\
Random Forest & 0.710 & 0.200 & $-0.300$ & 0.647 & 0.173 & $-0.238$ \\
\rowcolor{black!8}
Ridge         & \textbf{0.695} & \textbf{0.190} & \textbf{-0.280} & \textbf{0.621} & \textbf{0.165} & \textbf{-0.214} \\
\bottomrule
\end{tabular}
\end{table}

We consider the complete feature set to be $\mathcal{P}=\{1,\dots,p\}$ with $p=d-1$ features (columns of $\tilde Z_\ell$).
\emph{Reference} Ridge model is trained using all features $\mathcal{P}$ and we compute MDI importances $\{s_j\}_{j\in\mathcal{P}}$. By sorting $s_j$ in descending order, we get a ranking $\pi=(\pi_1,\pi_2,\dots,\pi_p)$ from most to least important.

We then perform greedy forward selection, a feature selection method based on the ranked order: for $k=1,\dots,p$, train the same ridge (same hyperparameters) on the top-$k$ features $\{\pi_1,\dots,\pi_k\}$ and evaluate its forecasting performance using the test block $\mathcal{T}^{\mathrm{test}}_\ell$ using nMAE. If $\mathrm{nMAE}_\ell(p)$ denotes nMAE with all $p$ features, we define the \emph{minimal feature count} as
\[
k_{\scriptstyle \ell}^{\star}=\min\{k\in\{1,\dots,p\}:\mathrm{nMAE}_{\scriptstyle \ell}(k)\le
1.05\,\mathrm{nMAE}_{\scriptstyle \ell}^{\mathrm{full}}\}
\]
i.e., the minimal number of features needed to achieve test performance (within a 5\% tolerance) that is comparable to performance with the complete feature set for lake $\ell$. In our experiment, \noindent we perform the ranking procedure \emph{separately} for each of the 30 lakes and average over all the lakes for the final ranking. We summarize across lakes via the distribution of $\{k_\ell^\star\}$ and report the corresponding feature subset(s) $\{\pi_1,\dots,\pi_{k_\ell^\star}\}$. 

\subsection{Joint selection of minimal observations--important feature}
\label{sec:joint-min-obs-feats}

\textbf{Here, we unify the sample-size and feature-selection strategies into a single decision procedure} that, for each lake, identifies the \emph{minimally sufficient configuration}: the smallest number of most-recent training observations $n$ and the fewest predictors $k$ that achieve near–full-history accuracy. This joint selection produces lake-level configurations that are then summarized across lakes to yield actionable guidance on “how much data” and “which measurements” are most essential. We call $(\hat n_\ell,\hat k_\ell)$ the lake’s \textit{minimal configuration}: the smallest $(n,k)$ meeting the accuracy tolerance.

\paragraph*{Decision for each lake}
We assume (i) a pre-test training pool of $N_{\mathrm{pre}}$ observations and a held-out test block (as defined in Sec.~\ref{sec:methods}); 
(ii) a per-lake ranking $\pi=(\pi_1,\dots,\pi_p)$ of the $p$ candidate predictors obtained in Sec.~\ref{sec:feature_ranking}; and 
(iii) the evaluation metric $\mathrm{nMAE}$ from Sec.~\ref{sec:methods}.

We train the ridge regression model on all $N_{\mathrm{pre}}$ observations using all $p$ predictors to obtain a reference error $\mathrm{nMAE}_{\text{full}}$ on the test block. 
A configuration $(n,k)$ is declared acceptable if it stays within a fixed tolerance of this reference:
\[
\tau \;=\; 1.05\,\mathrm{nMAE}_{\text{full}} \qquad\text{(5\% tolerance).}
\]

\paragraph*{Feasibility over $(n,k)$}
Over a grid $\mathcal{N}\subseteq\{1,\dots,N_{\mathrm{pre}}\}$ and $k\in\{1,\dots,p\}$, we train on the \emph{most-recent} $n$ pre-test observations using the top-$k$ predictors $\{\pi_1,\dots,\pi_k\}$ and compute $\mathrm{nMAE}(n,k)$ on the test block. 
We then define the \textbf{feasibility function} as,
\[
f(n,k) \;=\;
\begin{cases}
1, & \text{if } \mathrm{nMAE}(n,k)\le \tau \,\\[2pt]
0, & \text{otherwise},
\end{cases}
\]
where $\tau=1.05\,\mathrm{nMAE}_{\text{full}}$. 

Pairs with $n<k+1$ are excluded to prevent ill-posed or high-variance fits on tiny training subsets. 

\paragraph*{Minimally sufficient configuration}
Let $\mathcal{F}_\ell=\{(n,k): f(n,k)=1\}$ be the feasible set for lake $\ell$. We pick the \emph{lexicographically} smallest pair—i.e., minimize $n$ first and, for ties, minimize $k$:
\[
(\hat n_\ell,\hat k_\ell)=\operatorname*{lexmin}_{(n,k)\in\mathcal{F}_\ell} (n,k),
\]
where $\hat n_\ell$ denotes the selected length of the recent training history for lake $\ell$ and $\hat k_\ell$ denotes the selected number of predictors for that lake. If $\mathcal{F}_\ell=\varnothing$, we set $(\hat n_\ell,\hat k_\ell)=(N_{\mathrm{pre}},p)$,
i.e., the full-data, full-feature configuration. This returns the reference
model when no smaller subset meets the accuracy tolerance and ensures a well-defined output for every lake.

\paragraph*{Aggregated decision}
We compute $(\hat n_\ell,\hat k_\ell)$ for each lake and report the distributions of $\{\hat n_\ell\}$ and $\{\hat k_\ell\}$ via median and interquartile range (IQR). 

Compared to prior lake-forecasting studies, which typically vary training length or feature set in isolation, our feasibility function is a \emph{joint} acceptance test $f(n,k)$ tied to a lake-specific reference target, and a parameter-free lexicographic rule that returns a minimally sufficient configuration per lake. Additionally, for each lake, we record the predictor(s) selected by this joint rule and tabulate their frequencies across lakes.



\section{Results}
\label{sec:results}

\begin{figure}[!t]
  \centering
  \subfloat[Test $\mathrm{nMAE}$ (y-axis) versus the number of most-recent training samples $n$ (x-axis) for 30 lakes under the backward-expanding protocol. Each trace corresponds to one lake; the dashed vertical line marks the mean minimal sample count $\overline{n^\star}\!\approx\!176$.]{
    \includegraphics[width=0.96\linewidth]{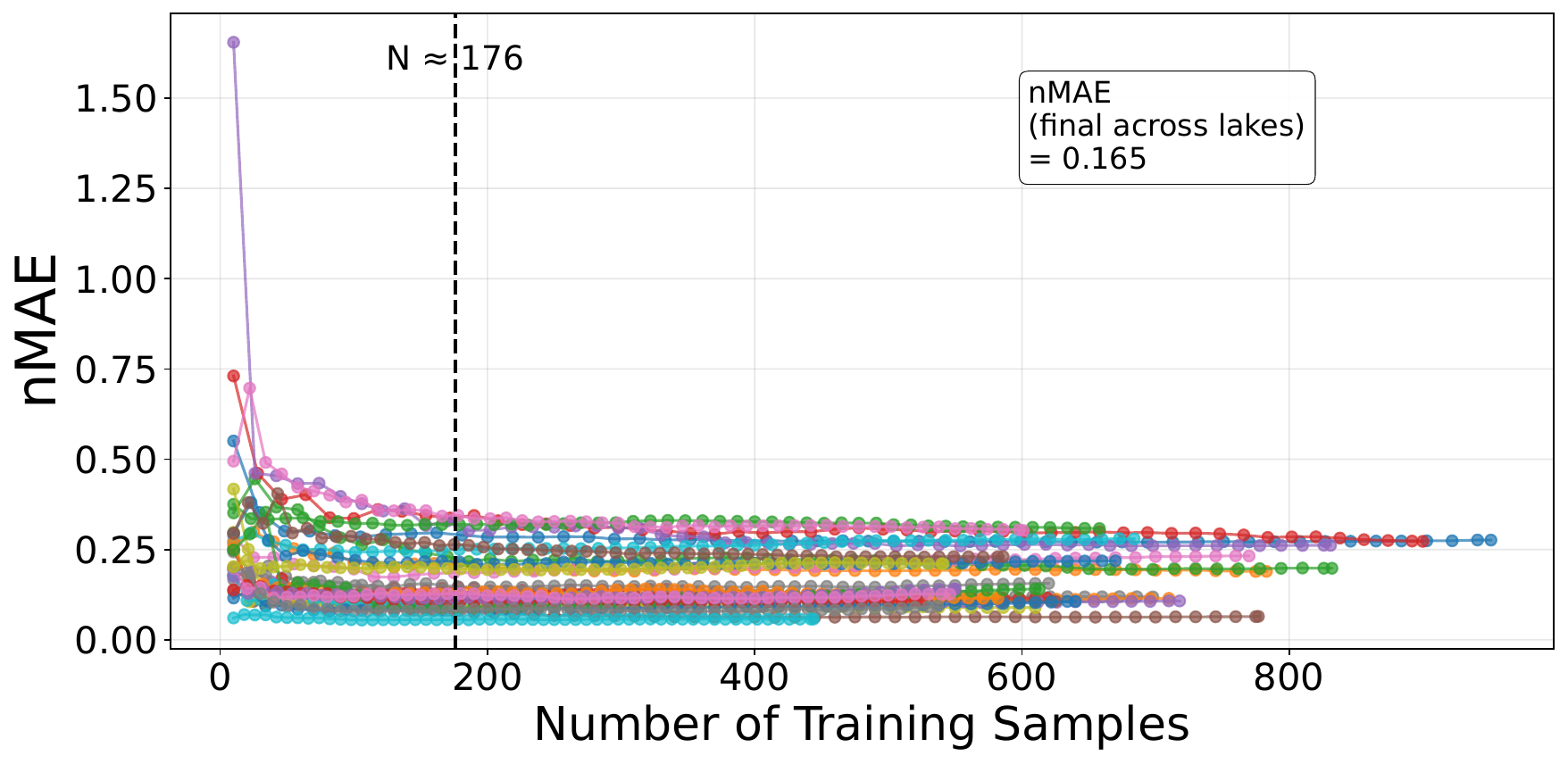}\label{fig:ridge-sampleset}
  }\\[2pt]
  \subfloat[Feature sufficiency via MDI ranking with ridge regression evaluation. The x-axis is the number of included predictors $k$ and the y-axis is mean test $\mathrm{nMAE}$. Each bar uses the top-$k$ predictors in the ranking. The highlighted four correspond to \texttt{TPEC}, \texttt{OXIC}, \texttt{SCHMIDT}, \texttt{CONDUCT}.]{
    \includegraphics[width=0.96\linewidth]{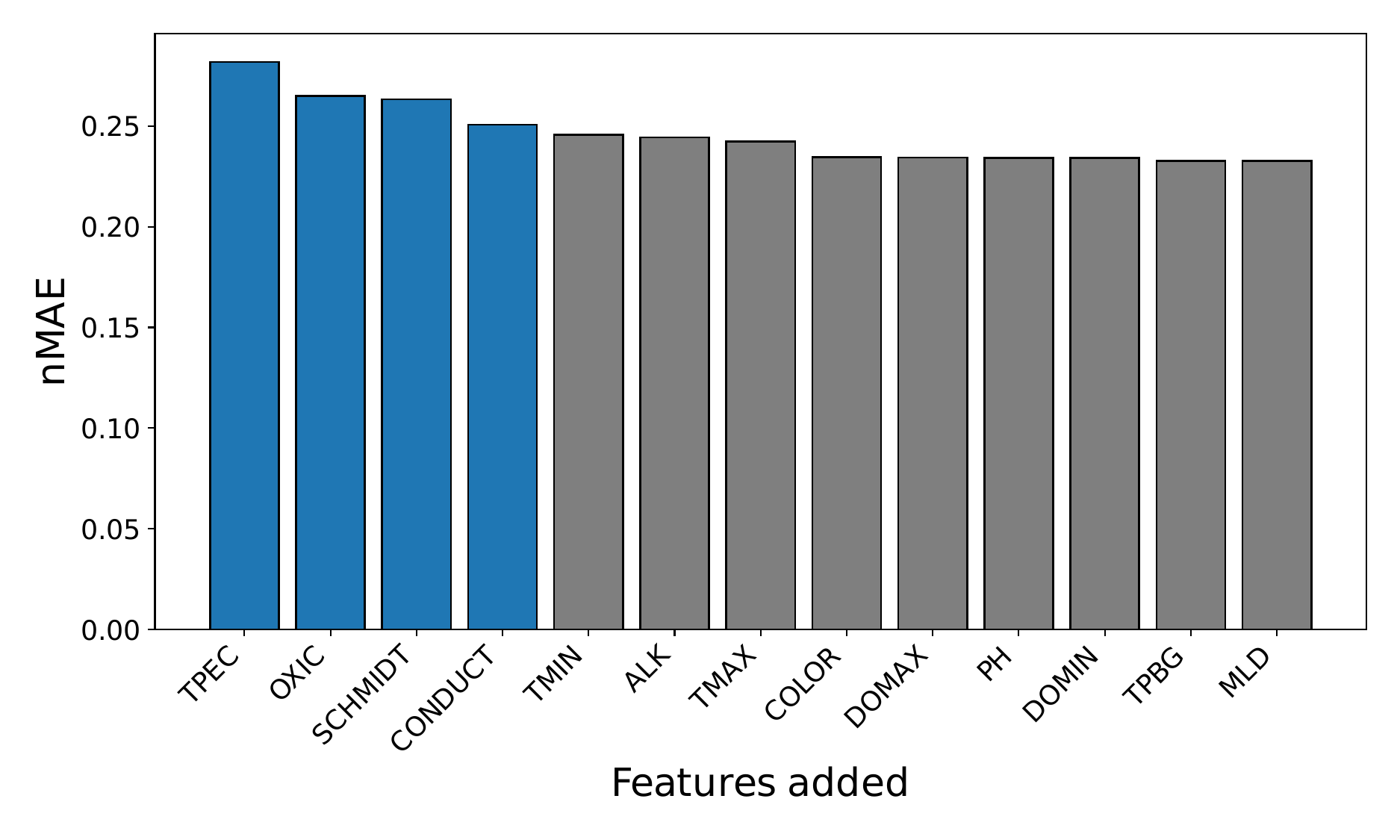}\label{fig:ridge-features}
  }
  \caption{(a) Minimal samples and (b) features for forecasting water quality (based on the 30 lakes).}
  \label{fig:ridge-sidebyside}
\end{figure}

\paragraph{\textbf{What is the minimal number of training samples needed?}}
\begin{sloppypar}
In Fig.~\ref{fig:ridge-sampleset}, we show how test nMAE changes with the number of training samples for each lake to capture the minimal number of samples needed. Notice that the error decreases rapidly over the first $\sim$50--100 samples and then flattens, indicating models reached a point of diminishing returns. The vertical dashed line denotes the (rounded) \emph{mean minimal sample count} across lakes, $\overline{n^\star}\!\approx\!176$, defined as the smallest $n$ for which a lake’s nMAE is within $5\%$ of the error received with the entire data. 
Beyond $\sim$150--200 samples, additional history yields limited improvement for most lakes with the ridge regression model.
\end{sloppypar}

\begin{figure}[!t]
  \centering
  \includegraphics[width=\linewidth]{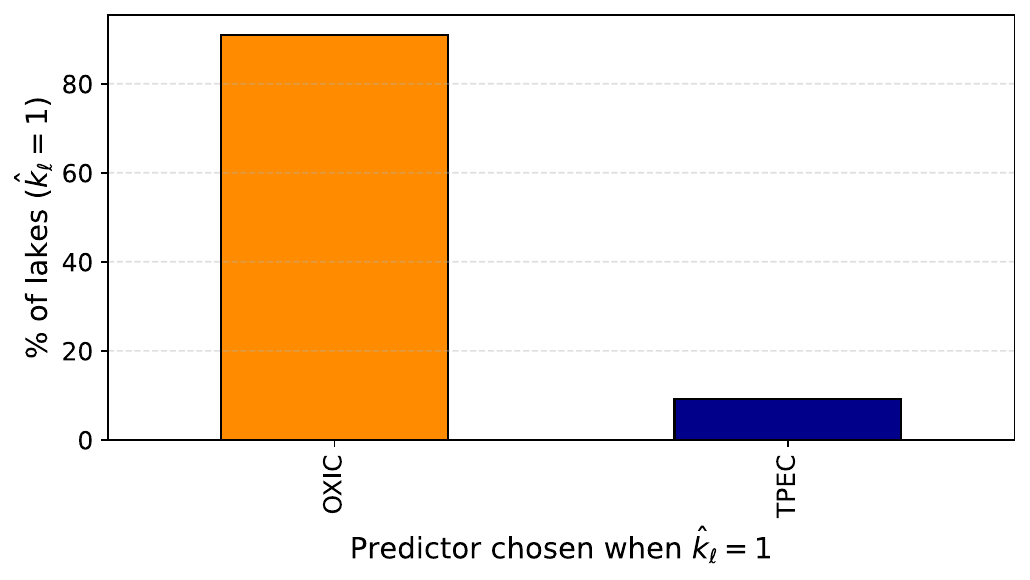}%
  \caption{Bar height (y-axis) gives the percentage of lakes selecting each predictor (x-axis) when the joint rule returns a single-feature model ($\hat{k}_\ell=1$). Under the 5\% tolerance, \texttt{OXIC} is selected for 90\% of lakes and \texttt{TPEC} for 10\%, indicating \texttt{OXIC} is the dominant lone predictor.}
  \label{fig:joint-feat}
\end{figure}

\paragraph{\textbf{How does nMAE enable depth-aware, cross-lake evaluation?}}

\noindent Sabattus Pond and Upper Narrows Pond from our dataset have similar test MAE (0.80\,m vs.\ 0.74\,m), which suggests comparable accuracy. However, their test-set mean SDDs differ widely (2.14\,m vs.\ 6.43\,m), suggesting that MAE is limited in capturing the distinction between shallow and deep lakes. Our custom evaluation metric, nMAE, after scale-normalizing MAE, reveals this when comparing MAE with nMAE: 0.37 for Sabattus (i.e., the average absolute error is 37\% of typical clarity) compared to 0.12 for Upper Narrows (12\%). Thus, for cross-lake differences, nMAE enables fair comparison across systems and lakes. 

\paragraph{\textbf{Which features are most important?}}

\(\{\texttt{TPEC},\ \texttt{OXIC},\ \texttt{SCHMIDT}, and \ \texttt{CONDUCT}\}\)- this subset of four features obtained nMAE within $5\%$ of the full-feature performance; adding further variables does not provide further performance gain beyond the 5\% that has been sacrificed. This reduced feature set provides a sampling-efficient model for scaling forecasting tools to data-poor, resource-limited regions and integrating field-deployable systems for climate-resilient water monitoring.

\paragraph{\textbf{Which joint minimal sample count-feature choice is best?}}

Applying the joint feasibility procedure to the 30-lake subset, we find that near–full-history accuracy (within 5\%) can typically be recovered with a short recent history and very few predictors: the minimally sufficient configuration has median $\hat n=\mathbf{64}$ observations with IQR $(\mathbf{129})$ and median $\hat k=\mathbf{1}$ with IQR $(\mathbf{1})$. Here, the median $\hat n$ means a typical lake needs about 64 of the most-recent samples to meet the 5\% target, while an IQR of 129 indicates that the middle 50\% of lakes span a window width of 129 samples around that median (reflecting heterogeneous sampling density across systems). Likewise, a median $\hat k=1$ with IQR 1 implies that at least half the lakes reach the target with a single predictor, and the middle half require at most one to two predictors.

Under the lexicographic rule we implemented, the most common joint outcome was a single-predictor model with a short recent history: when the rule returned $\hat k=1$ (20 lakes), \texttt{OXIC} was chosen in \textbf{90\%} of those lakes and \texttt{TPEC} in \textbf{10\%} (Fig.~\ref{fig:joint-feat}). Ten lakes had no feasible pair and reverted to the full-data/full-feature baseline. Hence, this minimal sample–feature configuration was optimal for our case. For lake scientists, such a strategy signals clear targets for recent-history length and a small, prioritized measurement set.

\section{Discussion}
\label{sec:discussion}

For most lakes, predictive performance improved rapidly with the first $\sim$50--100 observations and then exhibits diminishing returns. The estimated mean minimal sample count is ($\overline{n^\star}\!\approx\!176$. Fig.~\ref{fig:ridge-sampleset}) provides a practical target for monitoring programs: once this sample size is reached, further sampling can be decided based on the availability of resources. \texttt{TPEC}, \texttt{OXIC}, \texttt{SCHMIDT}, \texttt{CONDUCT}---these four features alone can provide reasonable forecasting performance compared to the full feature set, enabling compute-efficient deployment and streamlined data collection without sacrificing accuracy. These findings can inform government agencies and community science organizations developing cost-effective lake water monitoring strategies, particularly in regions where data collection is seasonal or capacity is limited.

\textbf{The outcome of the joint selection of minimal observations–features strategy}, $(\hat{n}_\ell,\hat{k}_\ell)$, turns directly into operational targets: maintain roughly $\hat{n}_\ell$ of the most-recent observations to stay within the 5\% accuracy tolerance, and prioritize only $\hat{k}_\ell$ measurements during routine visits. Programs can then reallocate effort accordingly: if the selected set emphasizes variables obtainable with in-situ sensors (e.g., temperature, conductivity), staff time shifts toward instrument deployment and maintenance with dense temporal coverage. If it emphasizes laboratory analytes (e.g., chlorophyll, total phosphorus), fieldwork can be scheduled as targeted campaigns at key seasons while supplementing with frequent, low-cost clarity checks by trained volunteers. By tying specific accuracy targets to the shortest recent history and the smallest required measurement set, this rule supports a monitoring-network design that explicitly balances accuracy, temporal coverage, and effort.

\textbf{Limitations.}
Our preliminary findings are based on several assumptions: nMAE (normalization by the test-set mean SDD) is used for evaluation, MICE imputation makes the Missing-At-Random assumption, and MDI feature importance can bias rankings toward features with high variance or many unique values. Our experiments use 30 lakes from the database and can be skewed towards lakes with longer records. We are actively expanding the study to incorporate additional lakes from the entire database. Additionally, tolerance and tie-break dependence: the 5\% accuracy tolerance and the lexicographic rule were design choices. Different settings (e.g., a cost-weighted tie-break) could change $(\hat n_\ell,\hat k_\ell)$.



\section{Conclusion}
\label{sec:conclusion}

This work contributes to scalable, ML-based approaches for early detection of water-quality degradation in freshwater systems, supporting climate resilience through data-driven environmental decision-making. 

\textit{Future works.}
For our future work, we plan to (a) incorporate physics-informed ML that encodes lake morphological information (e.g., depth) into the model; (b) evaluate the imputation quality by injecting artificial missingness; (c) investigate causal relationships among features and water quality (SDD); (d) conduct a sensitivity analysis of the accuracy tolerance and tie-breaking rule to quantify the stability of $(\hat n_\ell,\hat k_\ell)$; and (e) adopt a leakage-safe, nested train–validation–test evaluation in which imputation is fit on the training fold only and applied to validation-test, with model-threshold choices made on validation and final metrics reported on the held-out test set.

\section*{Acknowledgment}
We acknowledge D. Whitney King and Danielle Wain for helpful feedback and discussion on this work. We also thank the anonymous reviewers for their constructive feedback, which helped to improve this work. This work is supported by Colby College High Performance Computing Services and a Clare Boothe Luce Professorship from the Henry Luce Foundation.

\bibliographystyle{IEEEtran}
\bibliography{references} 

\appendices
\begin{table}[!t]
\caption{Per-lake training/test errors. The table also lists the 30 lakes used in the study.}
\label{tab:ridge-metrics}
\centering
\scriptsize
\begingroup
\setlength{\tabcolsep}{2pt}
\renewcommand{\arraystretch}{0.98}
\begin{tabular}{@{}lrrrr@{}}
\toprule
\textbf{Lake} & \textbf{Train MAE (m)} & \textbf{Test MAE (m)} & \textbf{Train nMAE} & \textbf{Test nMAE} \\
\midrule
Annabessacook Lake   & 1.091 & 1.223 & 0.304 & 0.283 \\
Highland Lake        & 0.836 & 0.940 & 0.165 & 0.183 \\
Cochnewagon Pond     & 1.060 & 1.013 & 0.200 & 0.163 \\
East Pond            & 1.012 & 1.167 & 0.259 & 0.222 \\
Sabattus Pond        & 0.605 & 0.542 & 0.294 & 0.253 \\
Woodbury Pond        & 0.608 & 0.421 & 0.093 & 0.064 \\
Cobbosseecontee Lake & 0.838 & 0.895 & 0.204 & 0.172 \\
Wesserunsett Lake    & 0.610 & 0.658 & 0.105 & 0.113 \\
Ingalls Pond         & 0.692 & 0.578 & 0.101 & 0.087 \\
Stearns Pond         & 0.560 & 0.551 & 0.109 & 0.108 \\
Elaine Pond & 0.949 & 1.142 & 0.228 & 0.216 \\
Mill Pond   & 1.195 & 1.075 & 0.318 & 0.260 \\
Wilson Pond          & 0.771 & 1.047 & 0.159 & 0.231 \\
Clary Lake           & 0.495 & 0.430 & 0.144 & 0.116 \\
Upper Narrows Pond   & 0.766 & 0.690 & 0.127 & 0.107 \\
Trickey Pond         & 1.062 & 0.968 & 0.106 & 0.103 \\
China Lake           & 0.757 & 1.079 & 0.219 & 0.263 \\
Taylor Pond          & 0.544 & 0.841 & 0.116 & 0.156 \\
Moose Pond           & 0.751 & 0.855 & 0.105 & 0.113 \\
Indian Pond          & 0.940 & 0.697 & 0.145 & 0.104 \\
Torsey Pond          & 0.606 & 0.679 & 0.101 & 0.103 \\
Meddybemps Lake      & 0.642 & 0.581 & 0.123 & 0.117 \\
Little Ossipee Lake  & 1.062 & 1.035 & 0.138 & 0.151 \\
Thomas Pond          & 0.696 & 0.739 & 0.108 & 0.114 \\
Wood Pond            & 0.563 & 0.705 & 0.088 & 0.114 \\
Keoka Lake           & 0.702 & 0.855 & 0.124 & 0.130 \\
Auburn Lake          & 1.141 & 1.612 & 0.161 & 0.238 \\
Long Pond            & 0.893 & 0.569 & 0.155 & 0.093 \\
Pushaw Lake          & 0.577 & 0.732 & 0.161 & 0.205 \\
Hancock Pond         & 0.520 & 0.379 & 0.071 & 0.053 \\
\bottomrule
\end{tabular}
\endgroup
\end{table}

\section{Train–Test Error Validation}\label{app:tt}

To verify that our training data support generalizable forecasts, we compare ridge errors on the training block and on the chronologically held-out test block(last 5 years). For each lake, we compute MAE and its nMAE on both the train and test sets. Similar train–test nMAE values indicate that the model learns signals from the training record and transfers them to the held-out period; large gaps would flag overfitting or inadequacies in the training data.

The results are summarized in Table~\ref{tab:ridge-metrics}. As the model is fit on the training block, we expect $\text{MAE}_{\text{train}}, \text{nMAE}_{\text{train}} \le \text{MAE}_{\text{test}}, \text{nMAE}_{\text{test}}$. This holds for most lakes, indicating adequate signal in the training data and reasonable generalization. A few lakes show \textbf{test $\le$ train} (Cochnewagon Pond, Woodbury Pond, Indian Pond, and Long Pond), which likely reflects higher variance within the training block.

\end{document}